\documentclass[10pt,conference,letterpaper]{IEEEtran}
\IEEEoverridecommandlockouts
\usepackage{cite}
\usepackage{amsmath,amssymb,amsfonts}
\usepackage{graphicx}
\usepackage{textcomp}
\usepackage{xcolor}
\def\BibTeX{{\rm B\kern-.05em{\sc i\kern-.025em b}\kern-.08em
    T\kern-.1667em\lower.7ex\hbox{E}\kern-.125emX}}

\usepackage{xcolor}
\usepackage{times}
\usepackage{epsfig}
\usepackage{graphicx}
\usepackage{amssymb}
\usepackage{xspace}
\usepackage{soul}
\usepackage{url}
\usepackage{tabularx}
\usepackage{adjustbox}
\usepackage{makecell}
\usepackage{hyphenat}
\usepackage{booktabs}
\usepackage{algorithmicx}
\usepackage{multirow}
\usepackage{footmisc}
\usepackage{setspace}
\usepackage{makecell}
\usepackage{authblk}
\usepackage[ruled,vlined,linesnumbered]{algorithm2e}

\SetCommentSty{mycommfont}
\SetKwInput{KwInput}{Input}                
\SetKwInput{KwOutput}{Output}              
\SetKwInput{KwRequire}{Require}            

\def \eg {\textit{e.g.}\xspace}
\def \ie {\textit{i.e.}\xspace}

\newcommand{\parahead}[1]{\vspace*{0.5ex}\noindent
	{\bfseries #1.}}

\newcommand{\systemname}{DTMM\xspace}
\newcommand{\fl}{filterlet\xspace}
\newcommand{\fls}{filterlets\xspace}
\newcommand{\flwcs}{FWCS\xspace}

\graphicspath{{Figures/}}

\begin{document}
\title{DTMM: Deploying TinyML Models on Extremely Weak IoT Devices with Pruning}

\author{Lixiang Han, Zhen Xiao, Zhenjiang Li \\ Department of Computer Science, City University of Hong Kong, China}

\maketitle

\begin{abstract}
DTMM is a library designed for efficient deployment and execution of machine learning models on weak IoT devices such as microcontroller units (MCUs). The motivation for designing DTMM comes from the emerging field of tiny machine learning (TinyML), which explores extending the reach of machine learning to many low-end IoT devices to achieve ubiquitous intelligence. Due to the weak capability of embedded devices, it is necessary to compress models by pruning enough weights before deploying. Although pruning has been studied extensively on many computing platforms, two key issues with pruning methods are exacerbated on MCUs: models need to be deeply compressed without significantly compromising accuracy, and they should perform efficiently after pruning. Current solutions only achieve one of these objectives, but not both. In this paper, we find that pruned models have great potential for efficient deployment and execution on MCUs. Therefore, we propose DTMM with pruning unit selection, pre-execution pruning optimizations, runtime acceleration, and post-execution low-cost storage to fill the gap for efficient deployment and execution of pruned models. It can be integrated into commercial ML frameworks for practical deployment, and a prototype system has been developed. Extensive experiments on various models show promising gains compared to state-of-the-art methods.
\end{abstract}


\section{Introduction}
\label{sec:intro}

Current Internet of Things (IoT) systems make wide use of resource-constrained devices such as microcontroller units (MCUs) for data acquisition in various sensor network applications in industry~\cite{duisterhof2019learning}, healthcare~\cite{fyntanidou2020iot,santagati2017implantable}, agriculture~\cite{jayaraman2016internet}, and smart city~\cite{roshan2021adaptive}, because MCUs have ultra-low power consumption, in milliwatts or microwatts~\cite{banbury2020benchmarking,huang2016battery}, which are suitable for sustainable and widespread deployment. Recent studies find that running machine learning (ML) models on such end devices can further provide useful device-side data processing to avoid frequent data transmissions (thereby reducing energy consumption), preserve data privacy and enable new application scenarios~\cite{fraternali2020ember}. Therefore, an emerging field of tiny machine learning (TinyML)~\cite{banbury2021micronets, David2021TensorFlowLM,warden2019tinyml} has appeared recently, focusing on adopting ML models on embedded IoT devices to expand their reach for ubiquitous intelligence.

However, the size and computation cost of ML models are usually not small, \eg, on the order of MB~\cite{he2016deep,simonyan2014very}, which contradicts to the extremely limited resource on MCUs. For example, a typical MCU (such as NUCLEO-F446RE) has only 512 KB of static random-access memory (SRAM) for temporary runtime data storage, 1 MB of on-chip embedded flash (eFlash) memory for program storage, and 180 MHz CPU frequency for task processing. Computation offloading is recently used to facilitate the deployment of ML models on MCUs to bypass their resource constraints~\cite{huang2022real}. However, offloading relies on additional infrastructure, such as edge for support, which is not conducive to large-scale deployment. To achieve fully autonomous operation and local inference, we revisit model compression and find that weight pruning~\cite{Li2017PruningFF,huang2020deepadapter} is still a viable but underexploited solution.

\begin{figure}
    \centering
    \includegraphics[width=0.4\textwidth]{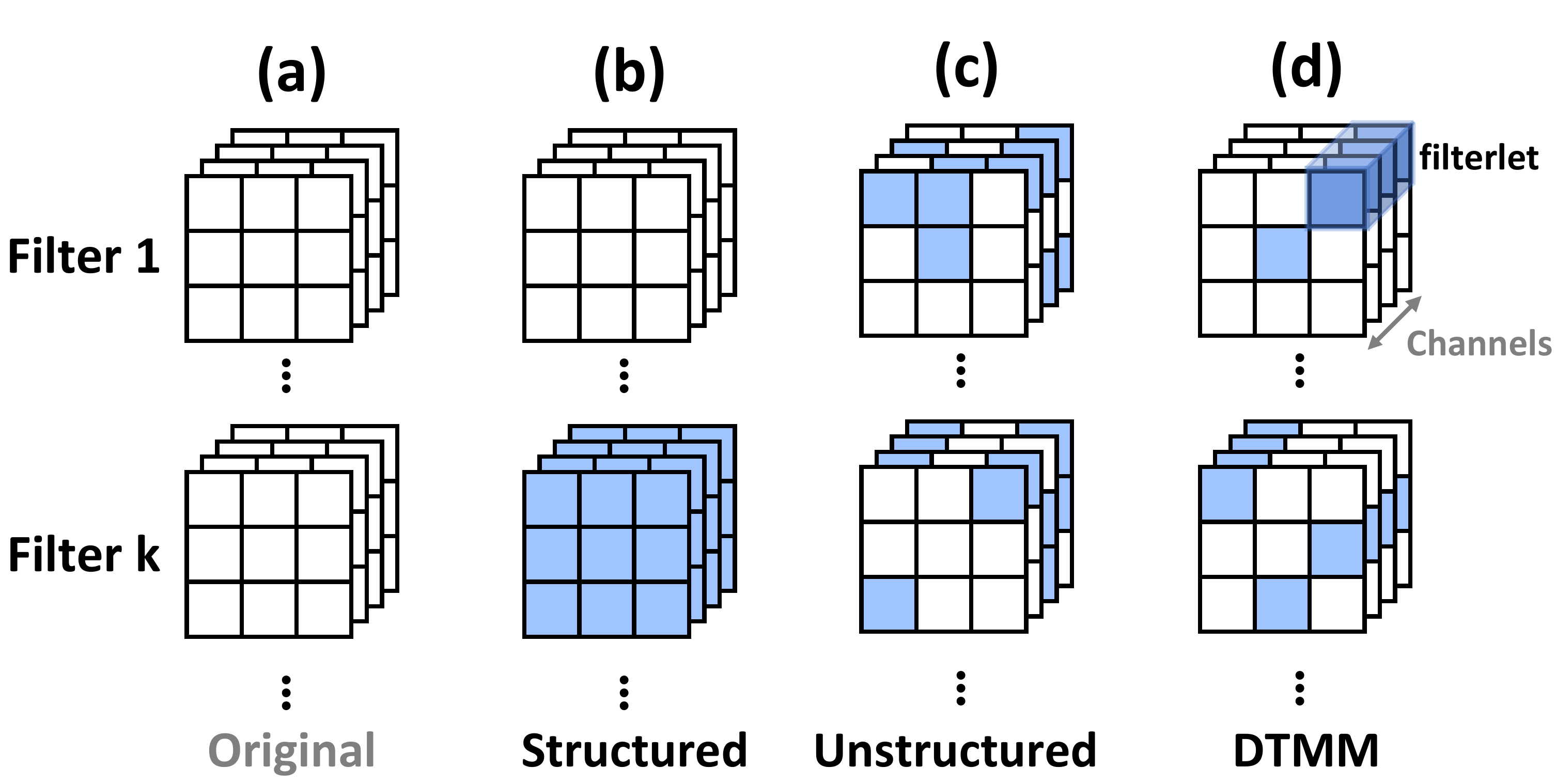}
    \vspace{-.1in}
    \caption{Illustration of weight pruning to fit TinyML models on MCUs. Blue squares represent the weights to be pruned. (a) One convolution layer contains multiple filters. (b) Structured pruning removes all the weights from the selected filter(s). (c) Unstructured pruning can remove arbitrary weights. (d) \systemname removes all the weights from the selected filterlets.}    
    \label{fig:pruning}
   \vspace{-.25in}
\end{figure}

Commercial ML frameworks, such as Tensorflow Lite Micro~\cite{David2021TensorFlowLM} and CMSIS-NN~\cite{Lai2018CMSISNNEN}, already supports \textit{structured} pruning~\cite{Li2017PruningFF,wen2016learning,sui2021chip}, which treats each filter (or other structures (\S\ref{sec:related})) as an atomic unit to remove all weights of selected filters to reduce the amount of model weights (Figure~\ref{fig:pruning}(b)). Structured pruning is easy to implement, but the pruned filters may still contain many useful weights, leading to coarse-grained pruning. The model accuracy drop significantly when high compression ratios are required on the MCU. 

To improve pruning granularity, another solution called \textit{unstructured pruning}~\cite{Han2016DeepCC} is proposed to prune arbitrary weights (Figure~\ref{fig:pruning}(c)). It achieves high accuracy by identifying and removing less important weights, but introduces two unique challenges on MCUs. First, due to the removal of arbitrary weights, each unpruned weight needs an index to record its existence, which incurs significant additional storage costs and lacks an efficient storage mechanism to fit the MCU's tight memory. Second, more importantly, the format of the remaining weights, indicated by their index values, is incompatible with commercial ML frameworks. It requires non-trivial overhead to convert these weights prior to inference~\cite{ma2021non}, which can significantly slow down model inference with low-end MCU processors (see \S\ref{sec:design:inf} for details).

To overcome these problems, we are inspired by recent attempts~\cite{wen2016learning,meng2020pruning} to group weights to form new pruning units with a granularity between entire filters and individual weights (\S\ref{sec:related}). We choose to group an entire line of weights at the same position across all channels in the filter as an atomic pruning unit for the MCU and name it \textit{filterlet}. Figure~\ref{fig:pruning}(d) shows that two and three filterlets (blue ones) are to be removed from Filter $1$ and $k$, respectively. Removing weights in the granularity of filterlets allows more flexible and finer-grained results than pruning filters, which thus can achieve higher compression ratio yet with little compromise of accuracy. 

\begin{table}[t]
\caption{Specification comparison between MCU and other platforms. ``RAS'' and ``N'' are short for Raspberry and NUCLEO.}
\vspace{-.1in}
\centering
\resizebox{\linewidth}{!}{
\begin{tabular}{l c c c c c c}
\hline
\hline
\textbf{Device}     &   \textbf{Freq.}  &   \textbf{RAM}    &   \textbf{Storage}    &   \textbf{Power}          \\
\hline
Jetson NX           &   1.1--1.9 GHz    &   8 GB            &   16 GB               &   20 W                    \\
Pixel 5             &   1.8--2.4 GHz    &   8 GB            &   128 GB              &   1.2 W    
\\
RAS Pi 3B           &   1.2 GHz         &   1 GB            &   Micro SD            &   1.3 W                   \\
N-F446RE            &   180 MHz         &   128 KB          &   512 KB              &   0.1 W          \\
\hline
\end{tabular}
}
\label{table:platforms}
\vspace{-.2in}
\end{table}

However, if we only change the pruning unit to filterlets, similar limitations to unstructured pruning still exist. The main reason we use this new unit is because we observe that weights in each filterlet are stored contiguously on the MCU (\S\ref{sec:design}). This local memory continuity can be used to design a new data structure to efficiently store pruned models and a novel operator to significantly speed up model inference, based on which we can also devise a specialized scheduler for deriving the optimal pruning strategy given resource constraints before actual pruning, resulting in a systematic and practical solution to deploy TinyML models on MCUs.

In this paper, we enable all these designs in \systemname, a full-stack library that fills in the important but missing pieces of post-execution low-cost storage, runtime inference acceleration and pre-execution pruning optimization to efficiently deploy TinyML models on MCUs through pruning. Pruning using filterlets  in \systemname is essentially a hybrid of the two types of pruning described above to leverage the advantages of both, but pruning is not the only step in deploying a model, and \systemname further avoids the respective shortcomings, making the pruned model finally usable and running on the real device.

We develop a prototype of \systemname, integrate it into TensorFlow Lite Micro~\cite{David2021TensorFlowLM} and evaluate its performance on Arm Corstone-300 of the Cortex-M55 processor~\cite{corstone} using various models, including VGG-11~\cite{simonyan2014very}, ResNet-12~\cite{he2016deep} and YOLO~\cite{redmon2016you} on three datasets: CIFAR-10~\cite{krizhevsky2009learning}, Visual Wake Words (VWW)~\cite{Chowdhery2019VisualWW}, and Face Detection Dataset Benchmark (FDDB)~\cite{fddbTech}. We compare \systemname with state-of-the-art structured and unstructured methods, CHIP~\cite{sui2021chip} and PatDNN~\cite{niu2020patdnn}. Overall, \systemname outperforms the structured method CHIP by reducing model size and inference latency up to 42.8\% and 27.7\%, respectively, without compromising accuracy. Compared to the unstructured method PatDNN, \systemname achieves up to 33.7\% and 74.6\% reduction on model size and inference latency, respectively, without compromising accuracy. In summary, this paper makes the following contributions:
\begin{itemize}
    \item We propose a novel library called \systemname for deploying TinyML models on weak IoT devices to achieve high compression ratio, high accuracy and small overhead.
    \item We choose a suitable pruning unit and design a set of new techniques for model storage, operator acceleration and optimization to realize the design of \systemname, which is compatible to commercial ML frameworks.
    \item We develop a prototype of \systemname and extensively evaluate its performance using different ML models and rich datasets, showing remarkable performance gains.
\end{itemize}


\section{Background}
\label{sec:pre}

\subsection{Computation Capacity of MCU}
\label{sec:pre:mcu}

MCU~\cite{mcus} is a small computer on a compact integrated circuit chip with CPU, memories, I/O peripherals, etc.

\textbf{1) Hardware footprint.} MCU is designed with low-end CPU (with 32--216 MHz operating frequencies, consuming only milliwatts or microwatts~\cite{banbury2020benchmarking}) and extremely small memories, such as SRAM (for temporary runtime data) in the range of 8--320 KB~\cite{mcus} and eFlash (for program and data storage) in the range of 64--1024 KB. The hardware footprint in the table shows that MCUs are typically small, low-cost and low-power. These advantages make MCUs economical and sustainable for widespread deployment in IoT applications~\cite{ roshan2021adaptive}.

\textbf{2) Comparison with other platforms.} In addition to weak devices, there are other popular computing platforms in IoT systems such as edge AI boards (\eg, Jetson NX)~\cite{nvdiajetson}, mobile devices (\eg, Pixel 5)~\cite{jia2022codl,RobuCIR-INFOCOM20}, and Raspberry Pi (\eg, Pi 3B)~\cite{pi2015raspberry}. Unlike the low cost of a few dollars for an MCU, these advanced platforms may cost hundreds of dollars to bring more computing resources, including a powerful CPU with GHz operating frequency and at least 1 GB memory. Table~\ref{table:platforms} summarizes the hardware specification differences between the MCU (\eg, NUCLEO-F446RE) and other computing platforms. In addition, these platforms are equipped with multi-core CPUs and GPUs. They can divide computing tasks into multiple threads and assign them to different cores to execute in parallel~\cite{wang2021asymo}, which are not supported on MCUs.

\subsection{Use of TinyML in IoT Applications}
\label{sec:pre:app}

In smart cities~\cite{xu2021limu}, MCUs are widely used in various sensors for data acquisition~\cite{zhang2017thoreau}. By running TinyML models locally, they can analyze sensory data and avoid frequent data transmissions (the main energy consumer~\cite{varshney2022judo}), which saves energy and helps extend the battery life. In tiny autonomous machines, ML models can also bring intelligence to small autonomous machines by performing more complex tasks, such as path planning and navigation in nano-drones~\cite{duisterhof2019learning}. In addition to saving energy, on-device processing can also preserve data privacy, such as health data on wristbands~\cite{fyntanidou2020iot} and user speech samples in keyword spotting~\cite{zhang2017hello}.

\subsection{Model Deployment Process on MCU}
\label{sec:pre:prune}

The process of deploying ML models with pruning on the MCU follows four main steps: 

\textit{Step 1)}: The pruner estimates the importance of each potential pruning unit (\eg, filterlets), which can be quantified by various metrics~\cite{Han2016DeepCC,lin2020dynamic}. The pruner then decides which weight units should be pruned based on the memory budget and performance requirements. This step in \systemname is handled by our pruning strategy scheduler in \S\ref{sec:design:sch}. 

\textit{Step 2)}: Given a pruning strategy, \ie, which weight units to prune, the pruner only resets them to zero in this step, because the model needs to be retrained (or fine-tuned). The retraining process only updates the unpruned weights.

\textit{Step 3)}: After retraining, the pruner applies the pruning strategy to remove weights and uses an appropriate data structure to store the pruned model with reduced model size. Quantization is also often used to further shrink the model due to tight memory and flash space on MCUs. This step in \systemname is covered by our storage structure design in \S\ref{sec:design:fl}.

\textit{Step 4)}: The pruned model can then be deployed on the MCU and operated by the ML framework for execution. Since commercial ML frameworks currently only support structured pruning, suitable runtime support is also required for deployment. A naive solution is to zero-pad the pruned weights to restore the structure of all filters, but this leads to unacceptable computational and storage overhead. Hence, we propose a novel operator design in \S\ref{sec:design:inf} that can run \systemname-pruned models directly to significantly speed up model inference.


\section{System Design}
\label{sec:design}

\begin{figure}[t]
    \centering
    \includegraphics[width=0.4\textwidth]{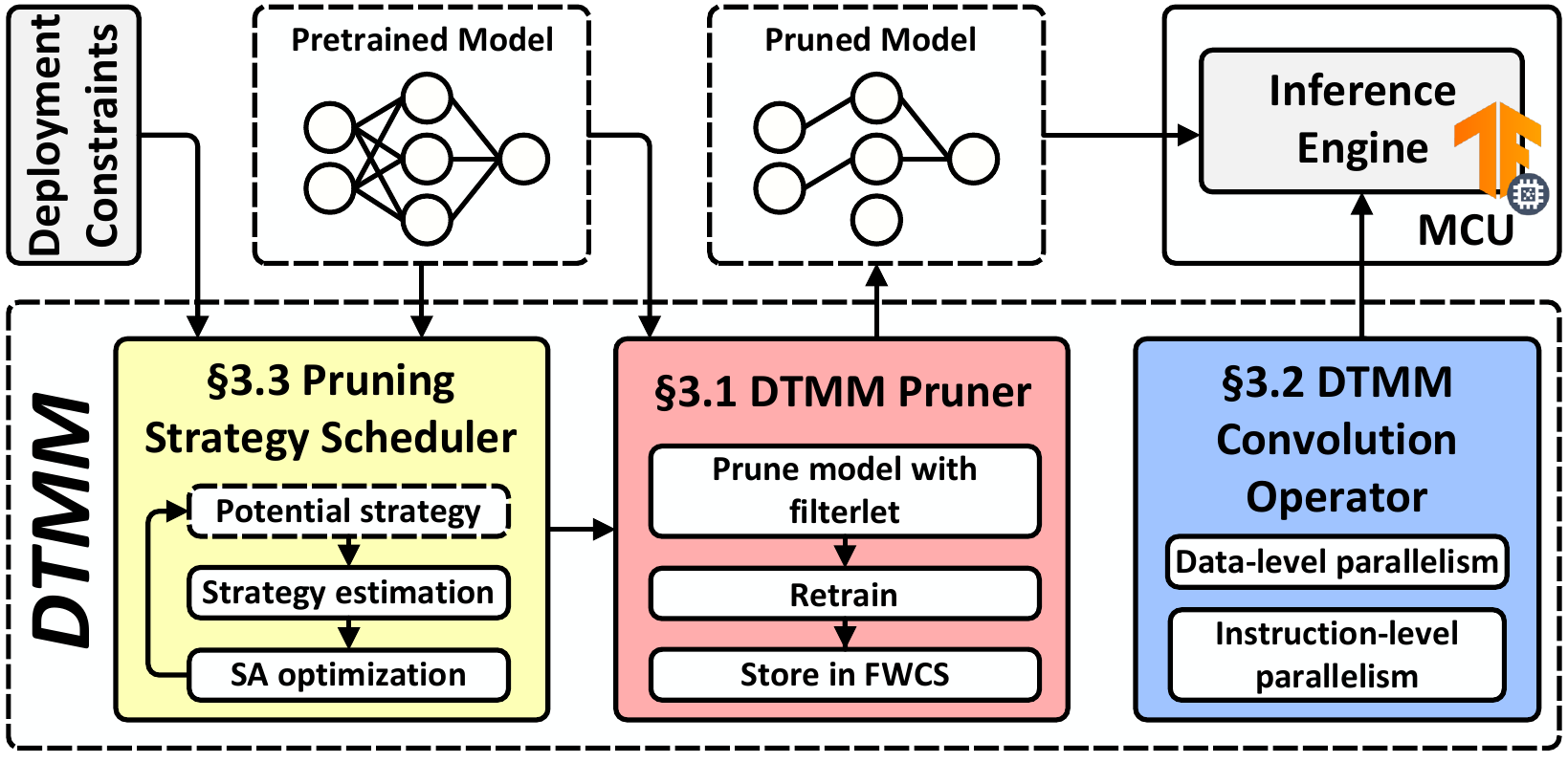}
    \vspace{-.1in}
    \caption{Overview of the \systemname design.}
    \vspace{-.2in}
    \label{fig:ptmm}
\end{figure}

The architecture of the \systemname design is illustrated in Figure~\ref{fig:ptmm} and consists of three main components:

\begin{itemize}
    \item \textbf{\systemname pruner (\S\ref{sec:design:fl})}: it prunes weights at a granularity of \fl and stores the remaining discrete weights with a compact data structure, called \flwcs, that is efficient for both storage and inference.

    \item \textbf{\systemname convolution operator (\S\ref{sec:design:inf})}: it enables unpruned weights stored in \flwcs to run in existing ML frameworks without reconstruction overhead, and accelerates the inference via single instruction, multiply data (SIMD) and instruction-level parallelism techniques.
    
    \item \textbf{Pruning strategy scheduler (\S\ref{sec:design:sch})}: it provides a pruning strategy for the pruner by formulating the search of the optimal strategy as an optimization problem to minimize the latency with accuracy and memory constraints.
\end{itemize}

\subsection{DTMM Pruner}
\label{sec:design:fl}

We first introduce the design of \systemname pruner, starting with a discussion of the problems of existing structured and unstructured pruning methods used on IoT devices.

\subsubsection{Problems in existing pruning methods}
\label{sec:design:fl:prob}

\begin{figure}[t]
    \centering
    \includegraphics[width=0.375\textwidth]{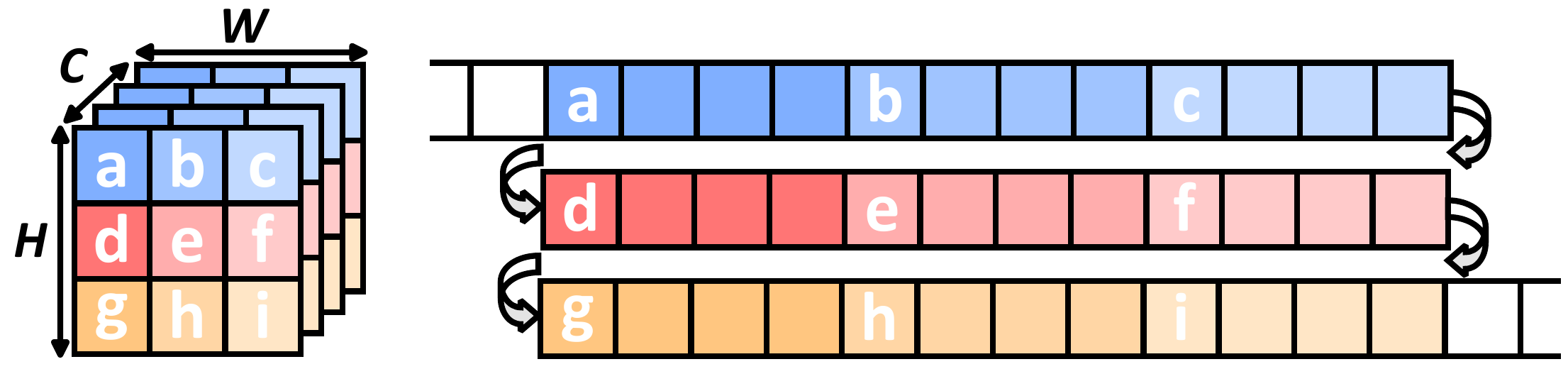}
    \vspace{-.1in}
    \caption{Weights in filters are stored contiguously in physical storage following a channel-major order. We plot the physical storage in three lines.}
    \label{fig:hwc}
    \vspace{-.2in}
\end{figure}

In a convolution layer, the weights are comprised of a set of \textit{filters}. Each filter contains $C$ \textit{kernels}, and one kernel has $H \times W$ weights. Different from other platforms, in commercial ML frameworks for MCUs, weights at the same position across all the channels are stored \textit{contiguously} in physical storage following a \textit{channel-major} order~\cite{Lai2018CMSISNNEN}, as shown in Figure~\ref{fig:hwc}.

\textit{Structured pruning}. When applying a pruning method, in addition to storing the values of the unpruned weights, their positions in the convolution layer should also be recorded to index the corresponding values from input feature maps during inference. The structured pruning methods~\cite{sui2021chip,Li2017PruningFF} remove weights at the granularity of filters, thus directly reducing model size. Meanwhile, it requires no additional index overhead since all computations are removed for each pruned filter. However, when the model needs to be deeply pruned for MCUs, coarse-grained pruning granularity (\ie, the entire filter) can affect the model accuracy.

\begin{figure}[h]
    \vspace{-.15in}
    \centering
    \includegraphics[width=0.45\textwidth]{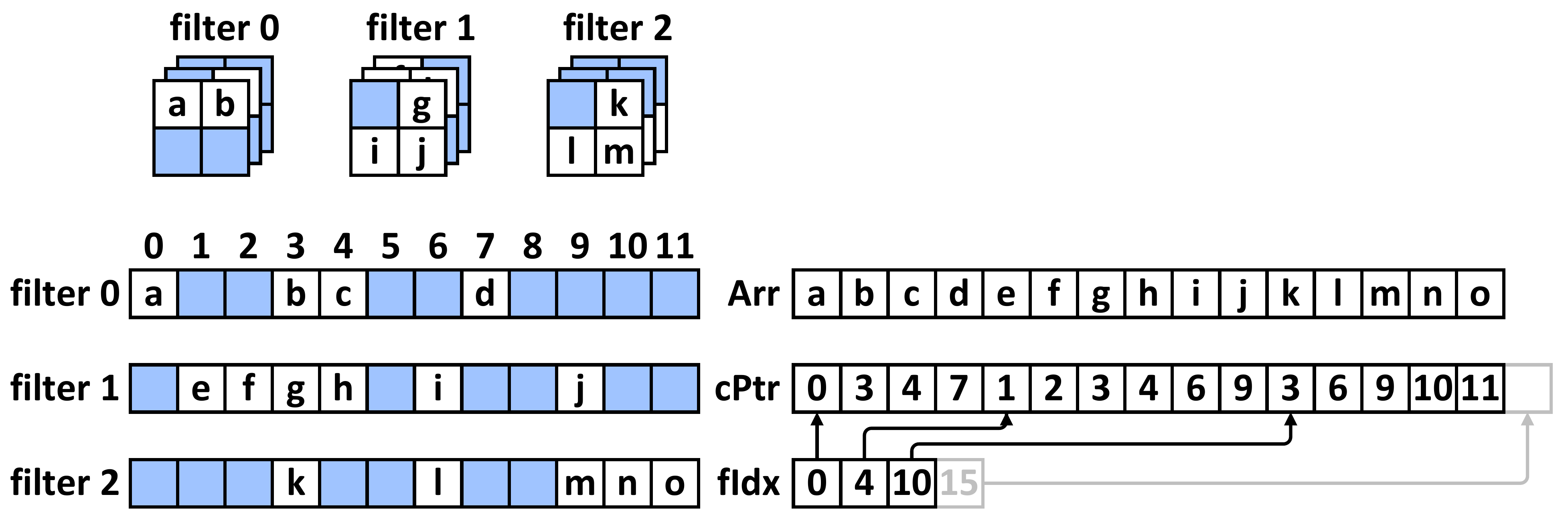}
    \vspace{-.1in}
    \caption{Illustration of how unpruned weights from three filters are managed with CSR structure in unstructured pruning.}
    \label{fig:fcsr}
    \vspace{-.1in}
\end{figure}

\textit{Unstructured pruning}. These methods~\cite{niu2020patdnn,Han2016DeepCC} preserve model accuracy by carefully choosing arbitrary weights to remove, but require extra storage to index the unpruned weights. Figure~\ref{fig:fcsr} shows a typical compressed sparse row (CSR) structure used in unstructured pruning~\cite{ma2021non}. Weights from different filters within the same convolution layer are stored consecutively. The values of the unpruned weights are stored in an array \texttt{Arr}, and their positions are represented using another two arrays \texttt{cPtr} and \texttt{fIdx}. \footnote{\texttt{cPtr} records the index of each unpruned weight in their corresponding filters. For \texttt{fIdx}, its length represents the amount of remaining filters. Each element in \texttt{fIdx} is a pointer pointing to the element in \texttt{cPtr}, which stores the position of the first unpruned element for the current filter.} Unstructured pruning introduces at least one index to record each unpruned weight, making the model size after pruning much larger than expected. More importantly, the format of unpruned weights, indicated by their index values, is not compatible with commercial ML frameworks. Recovering them to an executable format before inference requires non-trivial overhead, which can significantly slow down model inference on MCUs~\cite{ma2021non}.

Therefore, to prune the ML model with high accuracy, less storage and small execution overhead for MCUs, we next introduce the weight pruning with \fl and the new data structure design to efficiently store the unpruned weights.

\subsubsection{Weight pruning and storage in \systemname}
\label{sec:design:fl:prune}

\begin{figure}[t]
    \centering
    \includegraphics[width=0.45\textwidth]{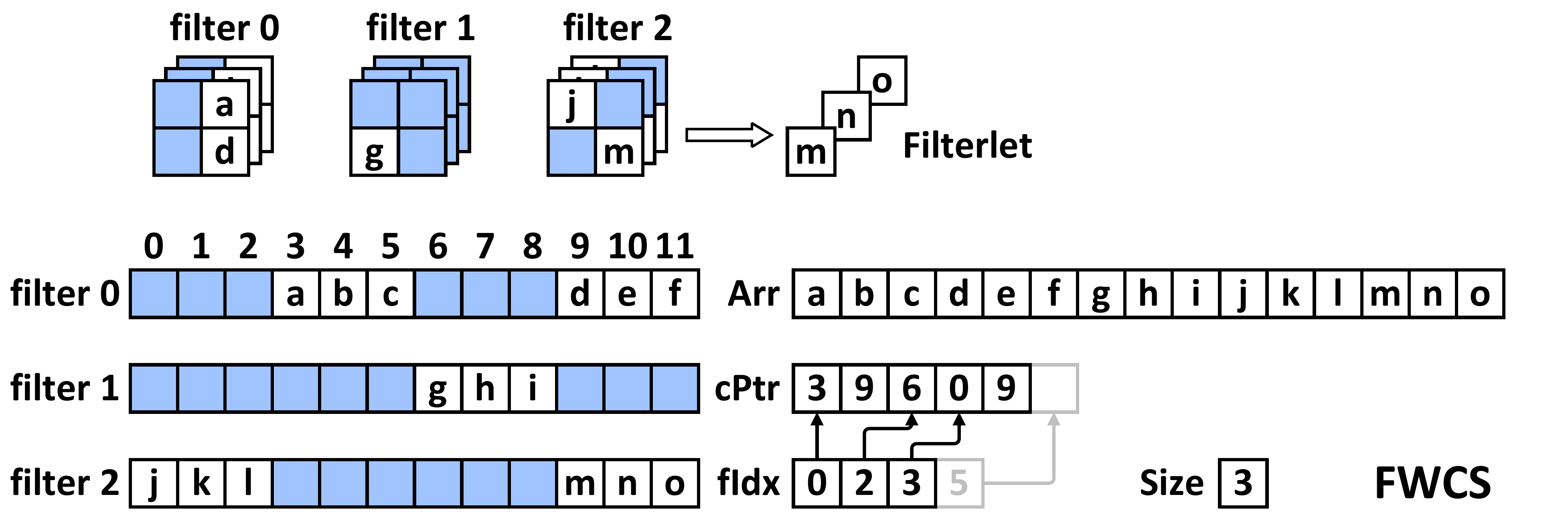}
    \vspace{-.1in}
    \caption{Illustration of how weights are pruned with \fl and how remaining weights are stored using \flwcs.}
    \label{fig:flet-wcs}
    \vspace{-.25 in}
\end{figure}

\parahead{Filterlet unit} For each filter, we group all the weights at the same position across all channels to form a \textit{\fl}. Figure~\ref{fig:flet-wcs} shows how to utilize \fl for pruning. In this example, each filter contains three channels, so a filterlet contains three weights, \eg, weights $\langle m,n,o \rangle$ represent a filterlet of ``filter 2''. Since weights are stored in a channel-major order on MCUs (Figure~\ref{fig:hwc}), all weights in each unpruned filterlet are located in contiguous space, which enables an efficient operator design for inference~(\S\ref{sec:design:inf}). Compared to structured pruning that regards the whole filter as an atomic unit to prune, \fl provides finer granularity to preserve important weights in each filter for better accuracy (\S\ref{sec:design:sch}).

\parahead{Compressed weight storage} Since each \fl is part of the filter, the weights from unpruned \fl also require extra storage to record their positions in the filter. To this end, we employ the key observation that all weights in a \fl are located in contiguous space to propose a compact data structure, called filterlet weight compressed storage (FWCS). Similar to the unstructured pruning, \flwcs stores the values of the remaining weights in an array \texttt{Arr}. The position information of each weight is encoded in other three arrays:

1) \texttt{size}: it stores the \fl size that is measured by the number of weights. For example, \texttt{size} is 3 in Figure~\ref{fig:flet-wcs}.

2) \texttt{cPtr}: it stores the index of the first weight of a \fl in the corresponding filter. Each element of \texttt{cPtr} corresponds to one \fl. For example, in Figure~\ref{fig:flet-wcs}, the second element of \texttt{cPtr} is 9, meaning that the index of the first weight $d$ in the second \fl ($\langle d,e,f \rangle$) is 9 in its filter.

3) \texttt{fIdx}: each of its element represents one filter in the current convolution layer. The value is a pointer to the index in \texttt{cPtr}, which stores the position of the first remaining \fl for the current filter. For example, in Figure~\ref{fig:flet-wcs}, the second element of \texttt{fIdx} represents the second filter (``filter 1''), and its value 2 means that the first \fl $\langle g,h,i \rangle$ of the second filter starts from column \texttt{cPtr[2]} (= 6).

\parahead{The effectiveness of \flwcs} Unlike unstructured pruning, we can store the position of each \fl instead of individual weights, leading to a substantial reduction in model size. With the number of channels $C$, removing the weight with \fl reduces the amounts of indexes by $C$ times. Hence, in Figure~\ref{fig:fcsr} and \ref{fig:flet-wcs}, pruning with \fl results in less storage overhead with the same number of pruned weights. 

In Figure~\ref{fig:eval_compression_format}, we conduct a preliminary experiment to compare the model size after pruning between unstructured pruning and \fl. We prune 90\% weights for five convolution layers with different configurations. Figure~\ref{fig:eval_compression_format}(a) shows that \flwcs can reduce the model size by 49.6\% on average compared to unstructured pruning. 

In addition to reducing model size, pruning with \fl also enables efficient model inference.  Direct execution of our pruned model in ML frameworks is slow (\S\ref{sec:design:inf}), as \flwcs lacks specialized runtime optimizations. To this end, we further design an efficient convolution operator to significantly speed up model inference. Next, we introduce this operator.

\subsection{DTMM Convolution Operator}
\label{sec:design:inf}

\subsubsection{Convolution operation at nutshell}
\label{sec:design:inf:basic}

A convolution layer performs a convolution operation on its filters and input feature maps. Each filter slides across the input feature maps, and the dot product between the filter and input feature maps is computed to generate the output feature map. Specifically, each value of the output feature map $o_{x, y, n}$ is computed by:
\begin{equation}
\label{eqn:con}
    o_{x,y,n} = \sum\nolimits_{h=0}^{H-1} \sum\nolimits_{w=0}^{W-1} \sum\nolimits_{c=0}^{C-1} (k_{h,w,c}^n \times f_{x+h,y+w,c}),\hspace{-2mm}
\end{equation} 
where $k_{h,w,c}^n$ and $f_{x+h,y+w,c}$ are the values of the $n$-th filter and input feature maps, respectively. 

\begin{figure}[t]
    \centering
    \includegraphics[width=0.45\textwidth]{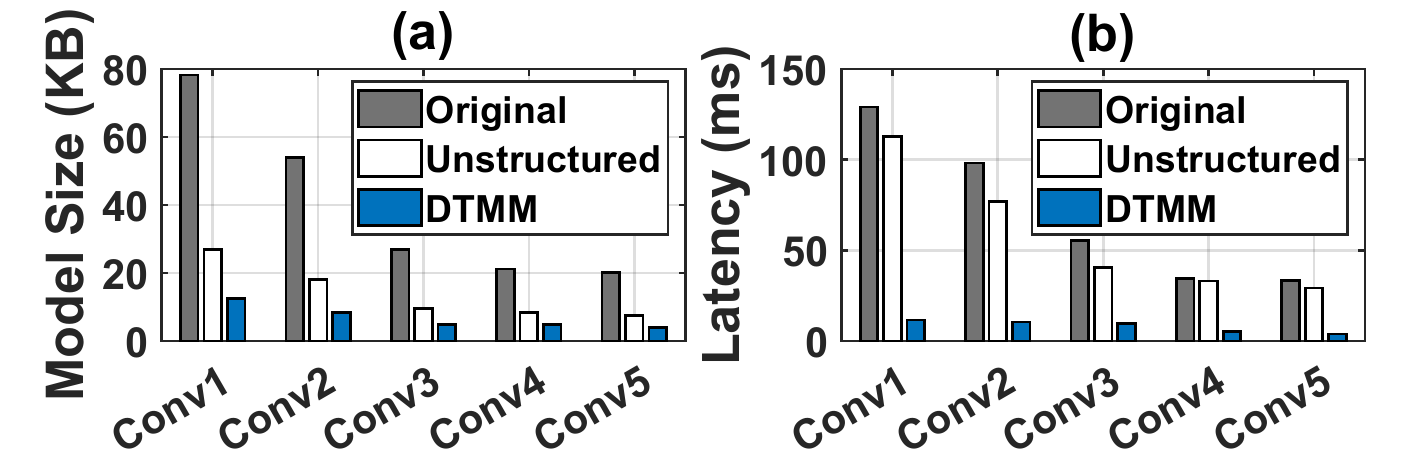}
    \vspace{-.1in}
    \caption{(a) Storage usage and (b) inference latency of convolution layers with 90\% weights pruned by different methods.}
    \label{fig:eval_compression_format}
    \vspace{-.2in}
\end{figure}

A naive way to compute Eq.~(\ref{eqn:con}) is to use nested loops, which is very slow. Therefore, ML frameworks for MCUs typically utilize the single instruction, multiple data (SIMD) technology to accelerate computation~\cite{Lai2018CMSISNNEN}. 
With SIMD, multiple operations in difference lanes (operands) are performed in parallel within an instruction achieving higher throughput.

The SIMD multiply-accumulate (MAC) operation is the main building block to implement dot product between the filter and input feature maps in a convolution layer. It computes and adds the dot product of two numbers to an accumulator ($a = a + \mathbf{b} \cdot \mathbf{c}$). SIMD then exploits the data-level parallelism of the hardware to increase the speed of computation based on the maximum lane number of the operands. 

\subsubsection{DTMM convolution with SIMD}

SIMD requires the weights in a filter to be contiguous in physical memory to efficiently load the weights and corresponding values in input feature maps for computation. However, due to the fine-grained weight removal, the unpruned weights of \systemname pruning (as well as unstructured pruning) become discrete, preventing a direct adoption of SIMD for speedup.

\parahead{1) Opportunity} Unlike individual weight removal in unstructured pruning, the weights of a \fl are contiguous in the physical memory since they come from the same kernel position across all channels. Unpruned weights in a convolution layer can be discrete, but the local memory continuity of weights in \fl brings further opportunities to exploit SIMD for inference speedup.

\begin{figure}[t]
    \centering
    \includegraphics[width=0.27\textwidth]{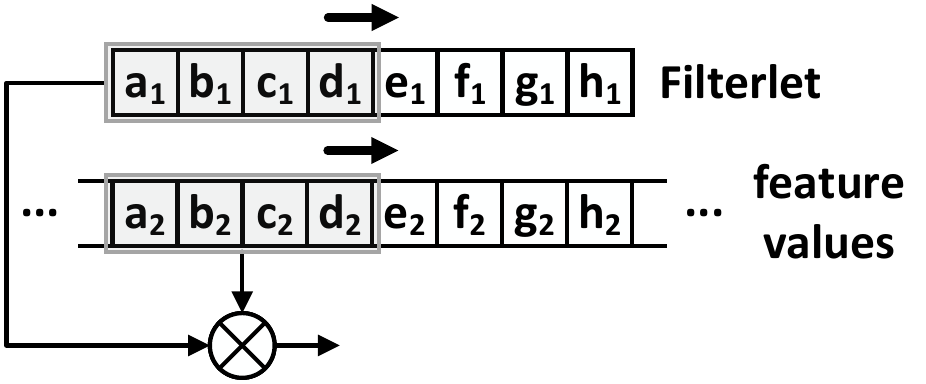}
    \vspace{-.1in}
    \caption{Weights in \fl can be operated by SIMD.}
    \label{fig:simd_mac}
    \vspace{-.2in}
\end{figure}

\parahead{2) When SIMD meets filterlet pruning} Figure~\ref{fig:simd_mac} illustrates how it works. In this example, a \fl has eight weights, and the lane number of the SIMD instruction is four. Since \fl weights are contiguous in physical memory, SIMD multiply-accumulate (MAC) can compute the dot product for four pairs of weights and feature map values in one instruction, instead of four. The speedup gain increases as the number of SIMD lanes increases, which is 2--16 in practice.

In general, \systemname convolution works as follows (Algorithm~\ref{alg:conv_basic}). The operator slides over the input feature map during convolution. For each step of sliding, a patch of the input feature map is prefetched, from which the operator extracts the values corresponding to the weights in \fl through the index information stored in \flwcs. The dot product between them is then performed using SIMD MAC operations.

\begin{algorithm}[h]
\caption{\label{alg:conv_basic}DTMM Convolution Operator}
\SetKwInOut{Input}{Input}\SetKwInOut{Output}{Output}
\begin{scriptsize}
\begin{spacing}{1.15}
\KwInput{Input feature maps: $I$; Filters in \flwcs format: $W$; Height, width and channel of output maps: $H_{out}$, $W_{out}$, $C_{out}$; Height and width of filters: $H_{ker}$ and $W_{ker}$; Convolution stride: $stride$; Number of SIMD lanes: $l$;}
\KwOutput{Output feature maps $O$;}
\SetAlgoVlined
\SetInd{0.05in}{0.1in}
\For{$h = 0$ \textbf{to} $H_{out} - 1$}{
    \For{$w = 0$ \textbf{to} $W_{out} - 1$}
    {
        $buf \gets \textnormal{Prefetch}(I, h \times stride, w \times stride, H_{ker}, W_{ker})$ \;
        \tcp{Fetch data necessary for $O_{h,w,c}$ computation}
        $i \gets 0$ \;
        \For{$c = 0$ \textbf{to} $C_{out} - 1$}{
            \For{$j = W.fIdx_c$ \textbf{to} $W.fIdx_{c+1} - 1$}{
                \For {$k = W.cPtr_j$ \textbf{to} $W.cPtr_{j} + W.size - 1$ \textbf{step} $l$}{
                    $O_{h,w,c} \gets buf_{k:k+l} \cdot W.arr_{i:i+l} + O_{h,w,c}$ \;

                    $i \gets i + l$\;
                }
            }
        }
    }
}
\end{spacing}
\end{scriptsize}
\end{algorithm}

\begin{figure}[h]
    \vspace{-.2in}
    \centering
    \includegraphics[width=0.4\textwidth]{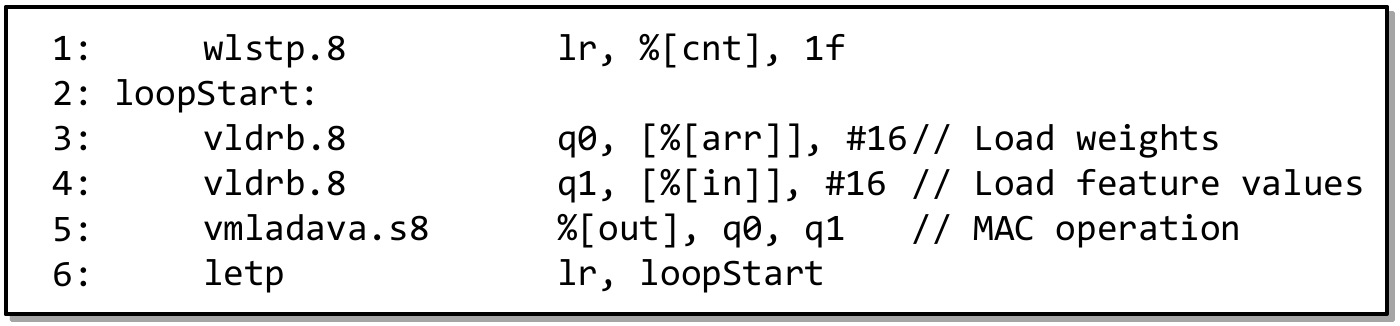}
    \vspace{-.1in}
    \caption{Main code block for our convolution operator with SIMD instructions on Armv8.1-M architecture.} 
    \label{fig:code_snip}
    \vspace{-.1in}
\end{figure}

We further transplant Algorithm~\ref{alg:conv_basic} into the ML framework, and Figure~\ref{fig:code_snip} shows the code block of our operator's main operation. In particular, SIMD load \texttt{vldrb.8} loads operand data into registers, \eg, \texttt{q0} and \texttt{q1}, and the SIMD MAC \texttt{vmladava.s8} computes dot product between the operand vectors. If the registers cannot hold a complete \fl, the computation is performed in iterations, and \texttt{wlstp.8} is used to record the number of iterations in \texttt{lr}. At the end of each iteration in line-6, \texttt{lr} is decremented by the number of lanes. If \texttt{lr} is non-zero, computation is repeated starting from line-2.

The speedup of our operator is mainly due to data-level parallelism, benefiting from the local weight continuity in a \fl to take advantage of SIMD. In addition to this data-level speedup, we also observe further acceleration opportunities through instruction-level parallelism in the next subsection.

\subsubsection{Instruction-level acceleration}
\label{sec:design:inf:reorganize}

When our convolution operator works, we observe that in Figure~\ref{fig:code_snip} there are two types of instructions executed by two different hardware components on the MCU, including
\begin{itemize}
    \item memory unit: it executes SIMD loads, and
    \item arithmetic logic unit (ALU): it executes SIMD MACs.
\end{itemize}

Since the memory unit and ALU are independent hardware components, there is an opportunity for these two types of instructions to be executed in parallel to further speed up computation, while we encounter the following problem.

\begin{figure}[t]
\centering
\includegraphics[width=0.35\textwidth]{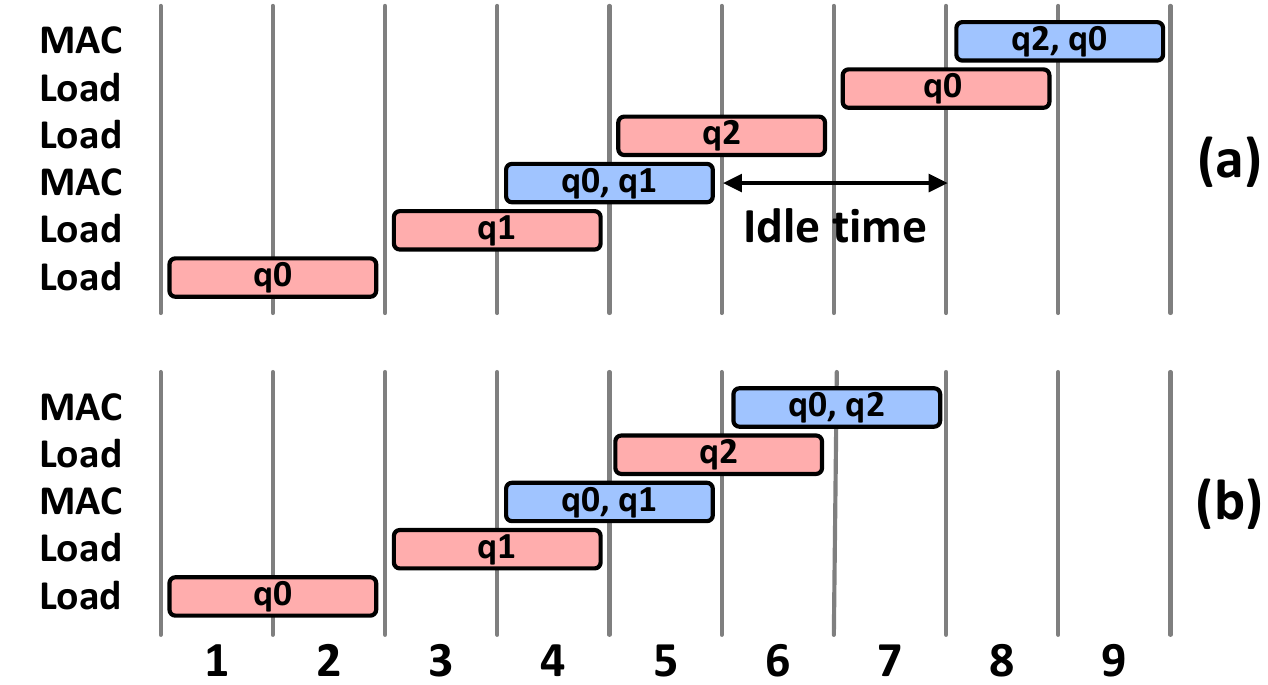}
\vspace{-.1in}
\caption{(a) Inefficient and (b) improved workflow of operations.}
\label{fig:pipline_original} 
\vspace{-.2in}
\end{figure}

\parahead{1) Problem} Figure~\ref{fig:pipline_original}(a) shows a snapshot of the memory unit (Load) and ALU (MAC) over time as we apply our operator to perform convolution. With recent vector processing technology~\cite{helium}, after partially loading data in the first CPU cycle, the ALU can immediately use it for computation, resulting in the overlap of Load and MAC, \eg, in cycle ``4''. 

We can see that in Figure~\ref{fig:pipline_original}(a), the ALU is idling during CPU cycles ``6'' and ``7'' as it waits for the memory unit to finish loading the two new operands into registers.\footnote{We introduce a new register \texttt{q2} in Figure~\ref{fig:pipline_original}(a) to improve execution efficiency. Executing our code in Figure~\ref{fig:code_snip} requires only two registers, but the next two new values are loaded in cycle ``6'' (to \texttt{q0}) and ``8'' (to \texttt{q1}), as the ALU is still using them in cycle ``5'', causing the ALU to wait three cycles before the next calculation.} We find that the ALU is periodically idle as the memory unit needs to be loaded with two new values for each MAC computation.

\parahead{2) Proposed solution} To avoid consecutive memory loads, we propose to maintain the value of a register, \eg, \texttt{q0}, and reorder the subsequent computations that require the value in \texttt{q0}. In this way, we only need to load one new value by the memory unit in the next period of time, so that the memory unit and the ALU can work alternatively without the ALU's idling waiting, as shown in Figure~\ref{fig:pipline_original}(b), where two MACs complete in seven CPU cycles vs.~nine in Figure~\ref{fig:pipline_original}(a).\footnote{We use two registers \texttt{q1} and \texttt{q2} that alternately load new values to maximize the efficiency of the ALU.}

In \systemname, we observe that the above idea applies to convolutions for the following reason. The default order of computation is to fix a position in the input feature map and iterate the dot product for all filters. However, the values in the feature map change every time because each unpruned \fl is in a different position in its filter. Thus, each MAC requires two memory loads, similar to Figure~\ref{fig:pipline_original} (a). 

However, if we adjust the order of computation by fixing the \fl weights in a register (\eg, \texttt{q0}), we can reuse their index information to alternately load values (at the same position but from different patches of input feature maps) into \texttt{q1} or \texttt{q2} and perform MAC (for \texttt{q0} and \texttt{q1}/\texttt{q2}). Thus, only one memory load is required per MAC before all computations requiring the current \fl weights are finished, similar to Figure~\ref{fig:pipline_original}(b). Note that the reordering only changes the computation sequence, while the result is unchanged.

\begin{figure}[t]
    \centering
    \includegraphics[width=0.4\textwidth]{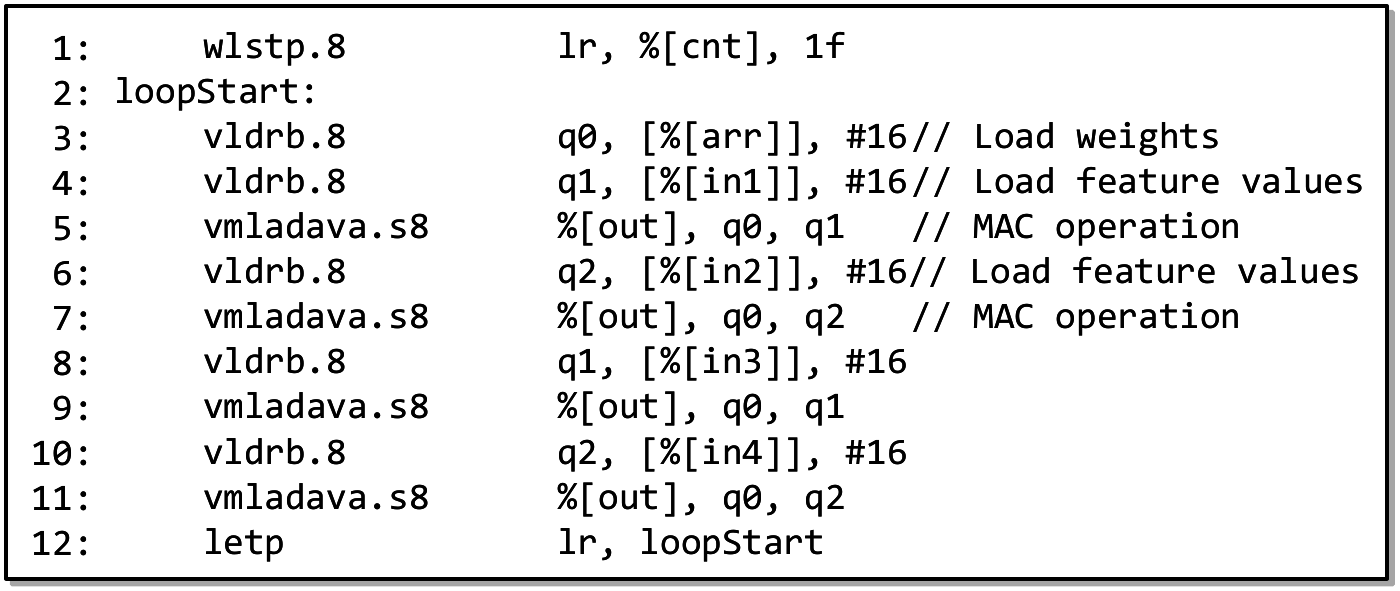}
    \vspace{-.1in}
    \caption{Improved convolution operator design using SIMD and instruction-level acceleration on Armv8.1-M.} 
    \label{fig:code_snip_two}
    \vspace{-.2 in}
\end{figure}

We incorporate this instruction-level acceleration to upgrade our convolution operator design in Figure~\ref{fig:code_snip_two}. Benefits come from two aspects. First, it avoids unnecessary idle states to improve CPU utilization. Second, it also avoids the redundant load of the same unpruned filterlet weights as they are used multiply times in the convolution. As shown in Figure~\ref{fig:eval_compression_format}(b), this design can reduce the inference latency by an average of 84.61\% compared to unstructured pruning.

\parahead{3) Summary} The key innovation of our acceleration design is not entirely in the use of SIMD and parallelism. The main challenge is that the chances of acceleration on MCUs without rich hardware features are very small. Therefore, we perform an in-depth analysis of convolutions to find a valuable opportunity to reshape their computational flow to fit the model structure and leverage (almost) the only advanced features (SIMD and parallelism) on MCUs to achieve acceleration.

\subsection{Pruning Strategy Scheduler}
\label{sec:design:sch}

A pruning strategy defines which pruning units should be removed from each layer. By making this strategy, one typical objective is to minimize the execution latency of the pruned model while ensuring that the accuracy is not compromised too much and model size becomes small enough. 

\subsubsection{Constraints of pruning strategy}

Before we provide an overall formulation of the pruning strategy, we need to first derive the following essential performance metrics to be used. 

\parahead{1) Accuracy} To avoid the impracticality of frequent evaluating the accuracy of pruned models during solving the pruning strategy problem, we employ a method inspired by previous works~\cite{Han2016DeepCC,lin2020dynamic} on estimating the importance of existing pruning units. We capture the change of loss after pruning individual units, and require the overall loss change for the pruned model to be less than a given $\Delta \mathcal{L}_{max}$, \eg, 0.001, to bypass deriving the accuracy of pruned models directly. Taylor expansion~\cite{Molchanov2017PruningCN} is used to measure the change of $\Delta \mathcal{L}$:
\begin{equation}
    \label{eql:taylor}
    \Delta \mathcal{L} \quad=\quad \left| \mathcal{L}(\mathcal{D}) - \mathcal{L}(\mathcal{D} \mid \mathbf{w}=\mathbf{0}) \right| \approx \left| \frac{\partial \mathcal{L}(\mathcal{D})}{\partial \mathbf{w}} \mathbf{w} \right|,
\end{equation}
where $\mathcal{D}$ is the training dataset and $\mathcal{L}(\mathcal{D} \mid \mathbf{w}=\mathbf{0})$ is the value of loss function when a set of weights $\mathbf{w}$ are pruned (set to zero). The gradient of the loss function to the weights can be obtained through the backpropagation algorithm. 

Now we compute $\Delta \mathcal{L}$ for each filterlet according to Eq.~(\ref{eql:taylor}) and use the error threshold $\Delta \mathcal{L}_{max}$ to obtain
the \fls that can be removed without exceeding $\Delta \mathcal{L}_{max}$. Then, we have the first constraint to ensure the model's performance after pruning:
\vspace{-.05in}
\begin{equation}
\label{eqn:acc}
\boldsymbol{C_{acc}}: \quad    \Delta \mathcal{L}(s) \leq \Delta \mathcal{L}_{max},
\vspace{-.08in}
\end{equation}
where $s$ is the decision variable $\langle \alpha_1, \cdots, \alpha_i, \cdots, \alpha_L \rangle$ and each $\alpha_i \in [0, 1]$ specifies the percentage of \fls to be removed from the $i$-th convolution layer.

\parahead{2) Model size and memory consumption} Model size for flash memory includes the following two parts in \systemname:
\begin{itemize}
    \item Amount of unpruned weights: $A_w = \sum\nolimits_{i=1}^{L} \alpha_i \times \left| \mathbf{K}_i \right|$, where $\left| \mathbf{K}_i \right|$ is the filter size in the $i$-th convolution layer and $L$ is the amount of convolution layers.
    \item Index overhead: $A_{idx} = m_0 \times \sum\nolimits_{i=1}^{L} (1 - \alpha_i) \times N_i \times H_i \times W_i$, where 
    $m_0$ is the bit width of each index (\eg, 16 bits), and the size of \texttt{rPtr} and \texttt{size} in \flwcs are omitted because they are very small.
\end{itemize}
With $m$-bit quantization, the model size is computed by:
\vspace{-.05in}
\begin{equation}
    {\rm Size}(s) \quad=\quad m \times A_w + A_{idx}.
    \vspace{-.08in}
\end{equation}

The memory consumption of SRAM to execute ML models is dominated by intermediate feature maps during inference. For \systemname pruned models, the output feature map size for the $i$-th convolution layer is: $M_i = \alpha_i \times N_i \times FH_i \times FW_i \times m$, where $FH_i$ and $FW_i$ are the height and width of the output feature map for the $i$-th convolution layer. During inference, models are executed layer by layer, and the intermediate feature maps are released after execution. Therefore, the memory usage is determined by the maximum memory consumption between any two adjacent convolution layers:
\vspace{-.05in}
\begin{equation}
    {\rm Ram}(s) \quad=\quad \mathop{max}\nolimits_{i}\{M_{i-1} + M_{i}\}, \quad i \in \left[1, L\right], 
    \vspace{-.08in}
\end{equation}
where $M_0$ is the size of the input data to the model.\footnote{For ML models with shortcuts, \eg, ResNet, the memory usage can be estimated from the topology of their computation graph~\cite{liberis2019neural}.} Then, we have the following two memory constraints:
\vspace{-.05in}
\begin{equation}
\label{eqn:sto}
\boldsymbol{C_{mem}}:~~~ {\rm Size}(s) \leq {\rm Mem}_{f};~{\rm Ram}(s) \leq {\rm Mem}_{r},
\vspace{-.05in}
\end{equation}
where ${\rm Mem}_{f}$ and ${\rm Mem}_{r}$ are the constraints of flash memory and SRAM, respectively.

\subsubsection{Pruning objective}

Within the above constraints, we aim to find the strategy that leads to the smallest execution latency.

\parahead{1) Execution latency per layer} We start with the execution latency $T_i$ of the single convolution operation in convolution layer $i$, which consists of three components:

\textit{i)} Feature fetching latency $T^{ft}_i$: the time used to fetch a patch of input feature values, and the volume is the same as a filter size. So, $T^{ft}_i = H_i \times W_i \times C_i \times t_{mem}$, where $t_{mem}$ is the time to fetch one value, $H_i$ and $W_i$ are the height and width of the kernel, and $C_i$ is the channel number in layer $i$.

\textit{ii)} Computation latency $T^{cm}_i$: the time used to first index the corresponding input feature values based on \flwcs, and then compute dot product via SIMD MAC for each unpruned \fl. It can be calculated by $T^{cm}_i = N_i \times H_i \times D_i \times (1 - \alpha_i) \times (t_{idx} + \lceil \frac{C_i}{l} \rceil \times t_{com})$, where $l$ is the maximum lane number supported by SIMD MAC, $N_i$ is the number of filters in convolution layer $i$, $\alpha_i$ is the percentage of filterlets to be pruned from this layer, $t_{idx}$ is the time to index a feature value, and $t_{com}$ is the time to perform a SIMD MAC.

\textit{iii)} Post-procesing latency $T^{ps}_i$: the post-processing time, $T^{ps}_i = N_i \times t_{post}$, where $t_{post}$ is the time of bias addition and quantization for each output value after the computation.

Therefore, we have $T_i = T^{ft}_i + T^{cm}_i + T^{ps}_i$, and the execution latency of the convolution layer $i$ is $\hat{T_i} = T_i \times FH_i \times FW_i$, where $FH_i$ and $FW_i$ are the height and width of the output feature map of layer $i$. Therefore, we propose to perform a regression on the parameters used in the calculations of $T^{ft}_i$, $T^{cm}_i$ and $T^{ps}_i$ as follows:
\vspace{-.05in}
\begin{equation}
    \label{eqn:regression}
    f(N_i, H_i ,W_i, C_i, \alpha_i) = \hat{T_i}.
    \vspace{-.05in}
\end{equation}

So far, the parameters $t_{mem}$, $t_{idx}$, $t_{com}$ and $t_{post}$ of execution latency model $f(\cdot)$ are undetermined. We can then use the measured latency of different configurations of convolution layers and the percentage of \fls to remove to train $f(\cdot)$ as a regression model. Based on our experiment, we find that utilizing 10 training samples can yield a low prediction error of $0.03$, as measured by the mean squared error.

\parahead{3) Pruning strategy} Finally, we formulate the search of the pruning strategy as an optimization problem to minimize the overall latency ${\rm Time}(s) = \sum_{i} \hat{T_i}$ over all the convolution layers. Given that convolution layers are the most time-consuming components, other layers (\eg, pooling) have relatively small impact on the overall model latency, which is not considered in $\rm Time(s)$. Therefore, we have:
\begin{equation}
    \min_{s}  \quad  {\rm Time}(s),
    \vspace{-.05in}
\end{equation}
subjected to $\boldsymbol{C_{acc}}$ in Eq.~(\ref{eqn:acc}) and $\boldsymbol{C_{mem}}$ in Eq.~(\ref{eqn:sto}). After decision variable $s$ is solved (\eg,  by simulated annealing (SA) solver), we obtain each $\alpha_i$ ($\in [0, 1]$), which specifies the percentage of filterlets to be removed from the $i$-th layer. 


\begin{figure*}[t]
\centering
\includegraphics[width=0.9\textwidth]{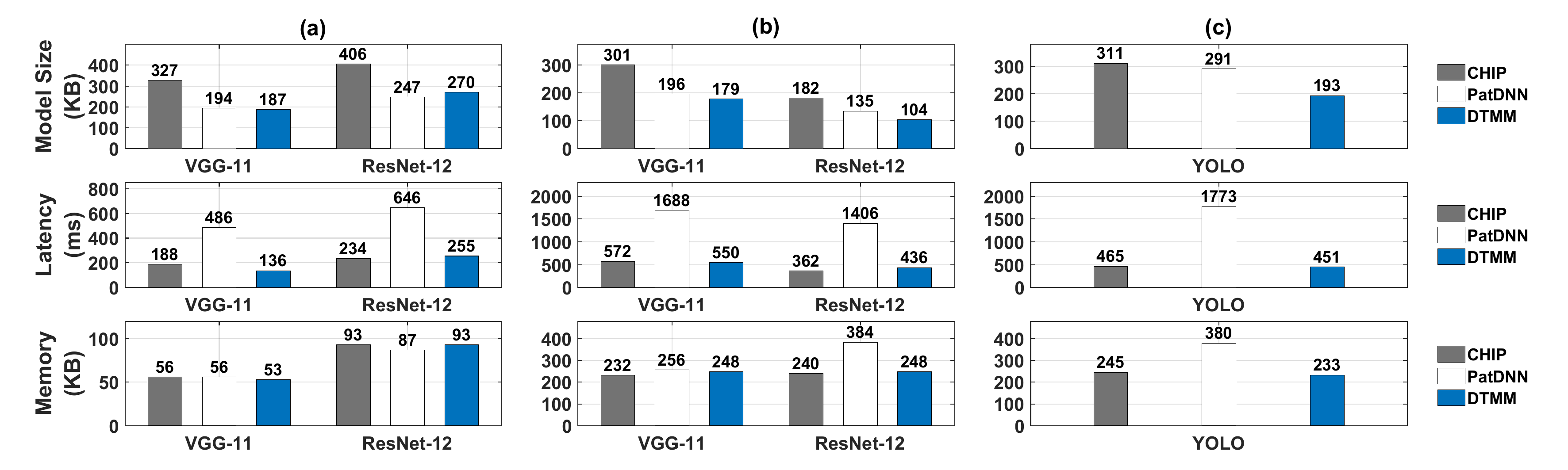}
\vspace{-.15in}
\caption{Overall performance of (a) VGG-11 and ResNet-12 models on CIFAR-10, (b) VGG-11 and ResNet-12 models on VWW, and (c) YOLO model on FDDB. The latency is measured when the processor operates at 120 MHz.}

\label{fig:eval_model_size} 
\vspace{-.25in}
\end{figure*}

\section{Implementation}

\label{sec:impl}

\parahead{1) Hardware} We implement \systemname with the Cortex-M55 processor~\cite{corstone}, the latest AI-capable processor with vector processing (\ie, Helium SIMD extension~\cite{helium}) for MCUs. The Cortex-M55 has eight 128-bit vector registers to store operands for SIMD instructions. Each vector register can be divided into lanes of width 8, 16 or 32 bits. For example, if each weight of a model is 8 bits (\ie, 8-bit quantization), a SIMD instruction can process 16 (=128/8) weights simultaneously. Each Helium vector instruction requires two CPU cycles. After the data is partially loaded into a register in the first cycle, the ALU can immediately use it for computations.

\parahead{2) Software} The software implementation of \systemname consists of two parts: offline toolchain and on-device runtime. We develop the offline toolchain using Python 3.6 and Tensorflow 2.6, and the on-device runtime using C and Helium SIMD extension in TensorFlow Lite Micro (TFLM)~\cite{David2021TensorFlowLM}. 

\textit{Offline toolchain}. The offline toolchain includes the scheduler that determines the pruning strategy, the pruner that performs filterlet-based pruning, and other modules such as quantization and retraining. To make our storage structure \flwcs used by the models after \systemname pruning compatible with TFLM, we further update TFLM's \texttt{schema.fbs} file, a format library (including metadata and runtime tensors), and add \flwcs as a new format member. 

\textit{On-device runtime}. We implement an on-device runtime based on TFLM. Our convolution operator is developed using C and inline Helium assembly. All these updates to TFLM do not affect its execution on other ML models. In particular, our convolution operator is launched only if a layer is pruned by \systemname; Otherwise, the traditional one is used.

\parahead{3) Models and datasets} We evaluate \systemname with three ML models widely adopted in the previous studies on MCUs, including VGG-11~\cite{simonyan2014very}, ResNet-12~\cite{he2016deep} and YOLO~\cite{redmon2016you}, in three popular ML tasks: image classification, visual wake words, and objection detection. For the classification tasks of image classification and visual wake words, we use VGG-11 and ResNet-12. For the object detection task, we use YOLO.

We train these ML models on the following datasets, which were used in the previous TinyML studies~\cite{kwon2022yono,lin2020mcunet}:
\begin{itemize}
    \item \textbf{1) CIFAR-10}~\cite{krizhevsky2009learning}: for image classification with 60 K of $32 \times 32$ RGB images of 10 different classes. 
    \item \textbf{2) Visual Wake Words (VWW)}~\cite{Chowdhery2019VisualWW}: for identifying the existence of person in an image. Images are resized to $64 \times 64 \times 3$ for more efficient execution on MCUs.
    \item \textbf{3) Face Detection Dataset Benchmark (FDDB)}~\cite{fddbTech}: for detecting the location and size of all faces in the image. Images are resized to $112 \times 112 \times 3$ for MCUs.
\end{itemize}

\renewcommand\arraystretch{1.2}
\begin{table}[t]
\caption{Particulars of each unpruned model with 8-bit quantization.}
\vspace{-.1in}
\centering
\resizebox{\linewidth}{!}{
\begin{tabular}{c|c|c|c|c}
\hline
\hline
\textbf{Dataset} & \textbf{Model} & \textbf{Accuracy} (\%) & \textbf{Model Size} (KB) & \textbf{Memory} (KB) \\
\hline
\multirow{2}{*}{\makecell{CIFAR-10\\(Image Classification)}} & VGG-11 & 90.19 & 1921 & 64\\
\cline{2-5}
& ResNet-12 & 90.15 & 1248 & 96\\
\hline
\multirow{2}{*}{\makecell{VWW\\(Visual Wake Words)}} & VGG-11 & 86.77 & 1919 & 256\\
\cline{2-5}
& ResNet-12 & 85.23 & 1246 & 384\\
\hline
\makecell*[c]{FDDB\\(Object Detection)} & \makecell*[c]{YOLO} & \makecell*[c]{60.29} & \makecell*[c]{2071} & \makecell*[c]{392} \\
\hline
\hline
\end{tabular}
}
\label{table:benchmarks}
\vspace{-0.2in}
\end{table}

Table~\ref{table:benchmarks} summarizes the accuracy, model size, and runtime memory for each unpruned model with 8-bit quantization.

\parahead{4) Training and pruning} We use the above three datasets to train a corresponding version of each model and then perform pruning, 8-bit quantization, and fine-turning on each version of the model before execution. For CIFAR-10, we follow the official training and testing division (5:1). We split the other two datasets (VWW and FDDB) into 70\% and 30\% for training/fine-tuning and testing, respectively.


\section{Evaluation}
\label{sec:eval}

\subsection{Overall performance}

\parahead{Methods} We compare the following pruning methods:

\textbf{1) CHIP}~\cite{sui2021chip}: a state-of-the-art structured pruning that detects and removes unimportant filters using the channel independence of the feature maps generated by each filter. 

\textbf{2) PatDNN}~\cite{niu2020patdnn}: a state-of-the-art unstructured pruning method. It includes an inference engine specifically for mobile devices. Since this engine is not MCU-compatible due to its reliance on the high-level parallel computing framework OpenCL, it is not included in the evaluation, and we implement a sparse convolution operator in TFLM for its inference.

\textbf{3) \systemname}: the method proposed in this paper.

\noindent \textbf{Performance comparison.} In this experiment, we compare the model size, execution latency, and runtime memory consumption of each model by maintaining model accuracy during pruning, \ie, with accuracy loss less than $0.5\%$. For the object detection task, accuracy is measured using the average precision (AP). The resource constraints for flash and SRAM are 512 KB and 256 KB, respectively.

\textit{1) Model size}. Figure~\ref{fig:eval_model_size} first shows the size of each model pruned by different methods. The unstructure-based PatDNN obtains a smaller model size than the structure-based method CHIP. \systemname can further reduce the model size due to its efficient storage design, outperforming CHIP and PatDNN by 39.53\% and 11.92\% on average, respectively.

\textit{2) Latency}. The second row in Figure~\ref{fig:eval_model_size} shows the latency of each model after pruning. Although PatDNN can achieve smaller mode sizes, its pruned models execute much slower than CHIP due to the complexity of handling discrete weights by the CSR structure during inference. Unlike PatDNN, \systemname also achieves small latency. Overall, the latency performance of \systemname outperforms CHIP and PatDNN by an average of 1.09\% and 68.70\%, respectively.

\textit{3) Runtime memory}. The runtime memory should fit within the SRAM of the device, and the runtime memory limit is set to 256KB in the evaluation. Both \systemname and CHIP can satisfy this constraint, but PatDNN may violate it in some cases due to the high indexing overhead.

\parahead{Accuracy as model size decreases} We prune more weights for each method and examine the resulting accuracy change of each pruned model. To this end, we relax the accuracy loss requirement in our pruning strategy scheduler to prune each model for a smaller size. We also prune each model to the same smaller size for other two methods. Due to the page limit, Figure~\ref{fig:eval_accuracy} (a--c) only plots the results of VGG-11 and ResNet-12 on CIFAR-10 and YOLO on FDDB. We can see that \systemname achieved the highest accuracy of pruned models than other two pruning methods (with similar model sizes) in most cases. The results are similar on other datasets.

\parahead{Analyze pruned models} We first analyze the percentage of weights pruned on each layer. Figure~\ref{fig:eval_accuracy}(d) shows the result of each layer of VGG-11 on CIFAR-10, \eg, 56.9\% of the weights are pruned from layer ``L1''. With \systemname, weights can be pruned selectively from each layer. Overall, 37.5--99.0\% of the weights from these layers are pruned.

We then analyze the components of each model pruned by \systemname. As shown in Figure~\ref{fig:eval_storage_usage}, 77.2--86.3\% of the model size after \systemname pruning is the weights. Indexing (and storage) overhead and other overhead (\eg, quantization parameters and metadata) are only 1.1--3.2\% and 12.6--13.9\%, respectively, which shows the effectiveness of our FWCS structure design. In contrast, for similar model sizes, the useful weights of each model pruned by PatDNN only count as 27.9--28.4\%, whereas its indexing and storage overhead is large, which explains why the model pruned by PatDNN loses more accuracy.

\vspace{-.01in}
\subsection{Micro-benchmarks}

\begin{figure*}[t]
    \begin{minipage}[t]{0.32\linewidth}
        \centering
        \includegraphics[width=1.0\textwidth]{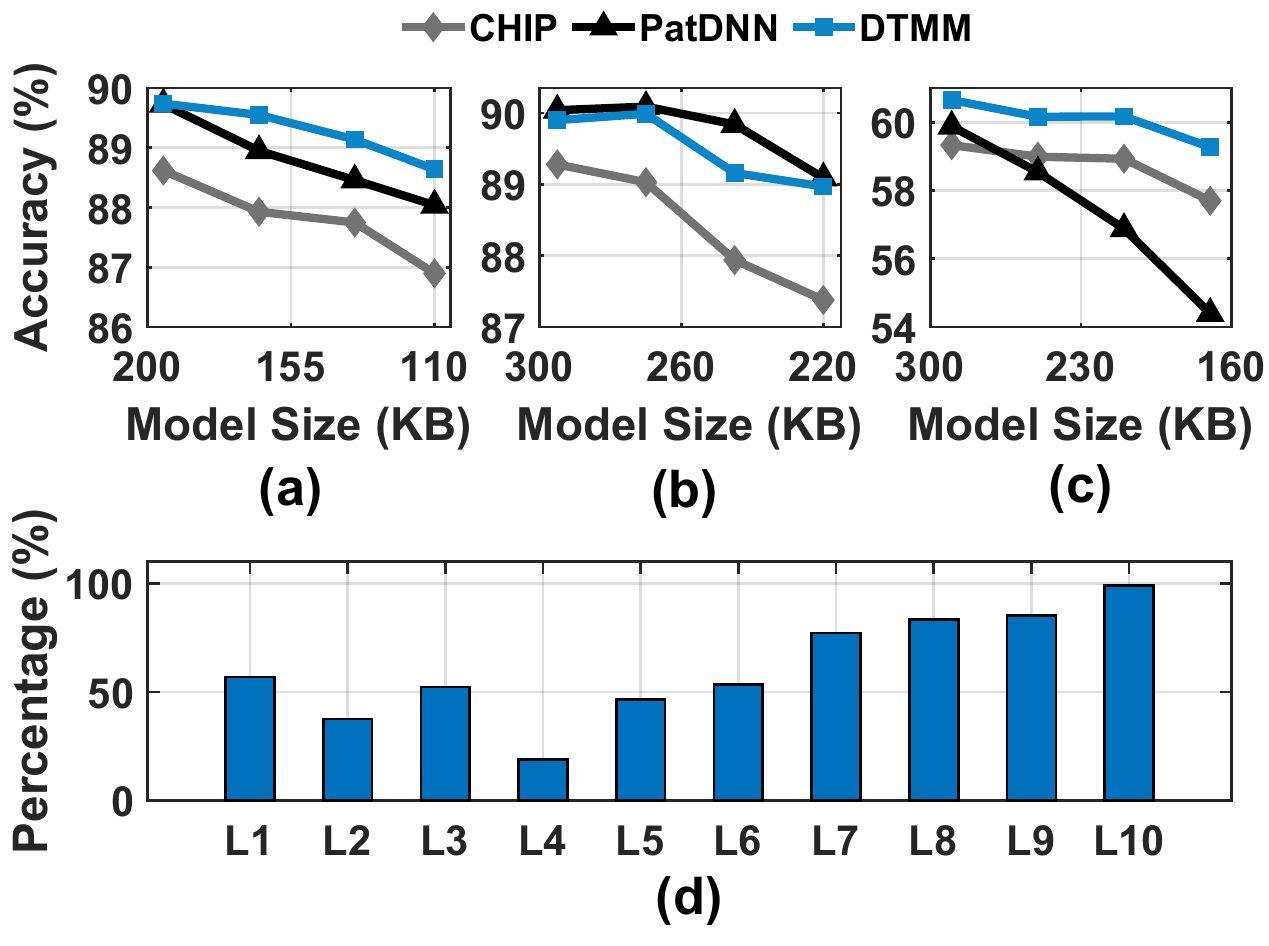}
        \vspace{-.25in}
        \caption{Accuracy when (a) VGG-11, (b) ResNet-12, and (c) YOLO are pruned to smaller sizes. (d) Percentages of VGG-11 weights pruned.}
        \label{fig:eval_accuracy} 
    \end{minipage}
    \hspace{0.1cm}
    \begin{minipage}[t]{0.32\linewidth}
        \centering
        \includegraphics[width=1.0\textwidth]{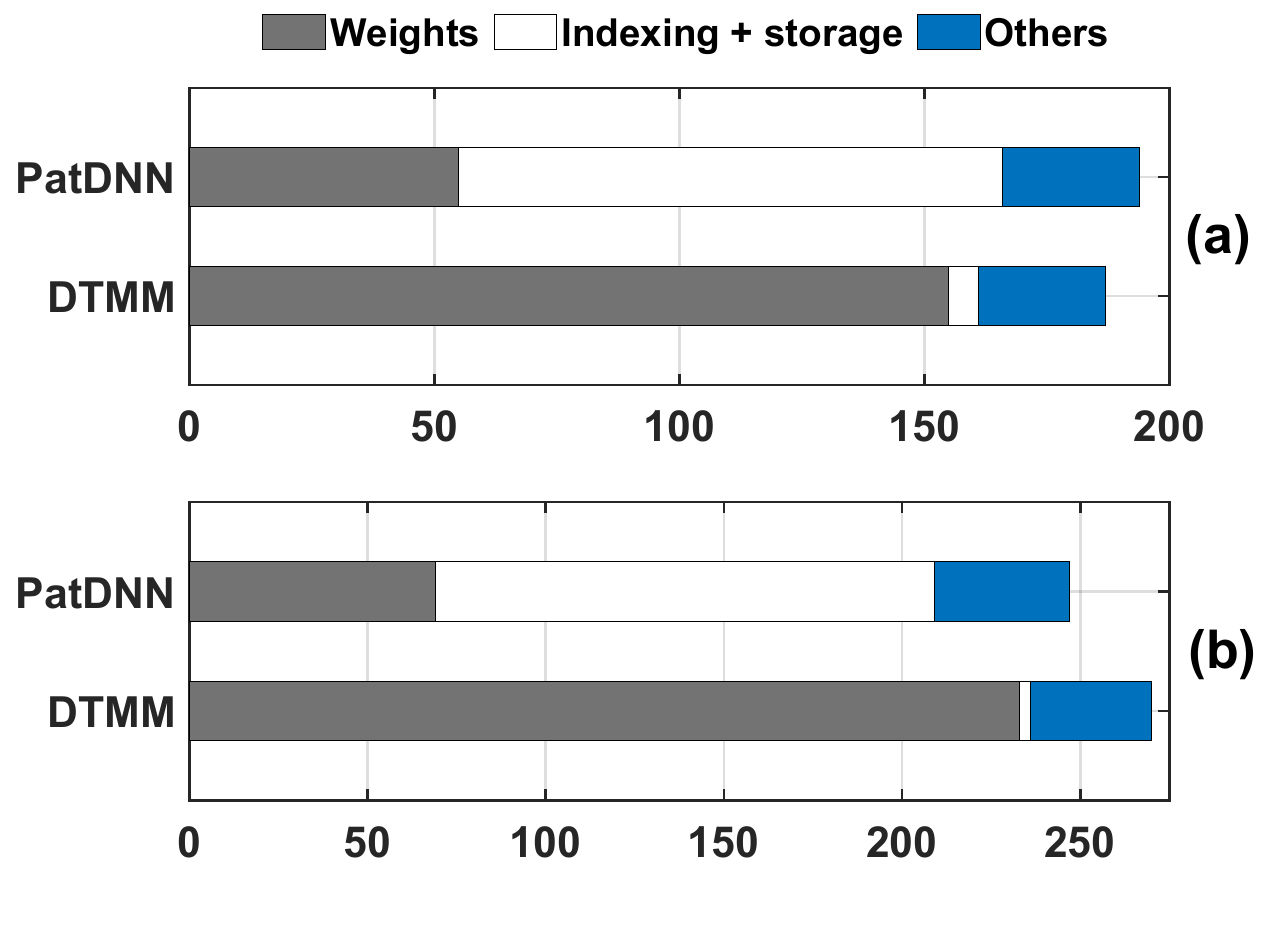}
        \vspace{-.25in}
        \caption{Model size breakdowns for (a) VGG-11 and (b) ResNet-12 when they are pruned by PatDNN and \systemname on the CIFAR-10 dataset.}
        \label{fig:eval_storage_usage} 
    \end{minipage}
    \hspace{0.1cm}
    \begin{minipage}[t]{0.32\linewidth}
    \centering
    \includegraphics[width=1.0\textwidth]{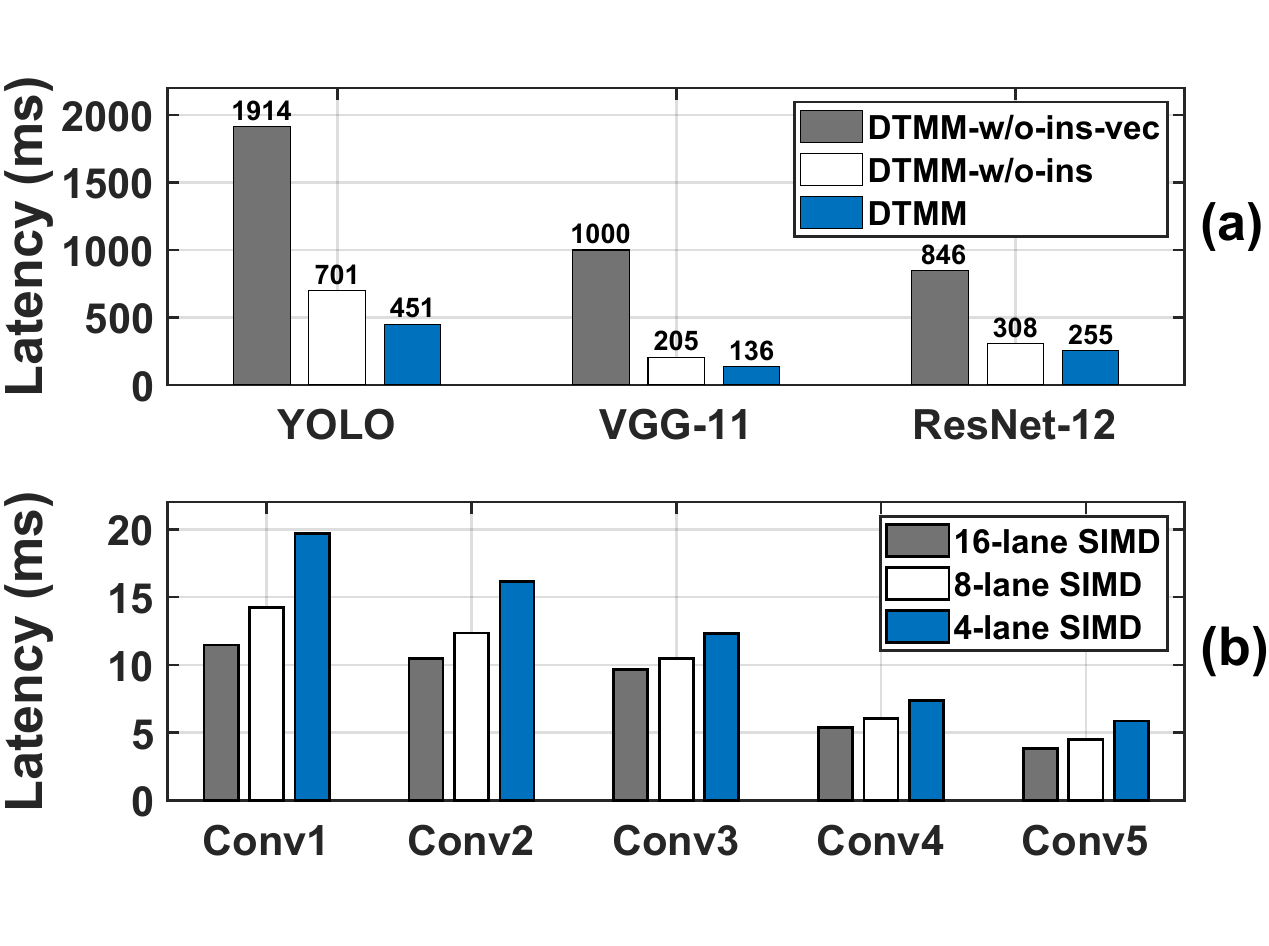}
    \vspace{-.25in}
    \caption{\label{fig:eval_operator_ablation} (a) Ablation study with three \systemname versions. (b) Inference latency of convolution layers when the number of SIMD lanes changes.}
    \end{minipage}
    \vspace{-.2in}
\end{figure*}

\parahead{Ablation Study} We first conduct an ablation study to examine the efficacy of two designs in our convolution operator. To this end, we develop two intermediate versions of \systemname:
\begin{itemize}
    \item \textbf{``DTMM-w/o-ins''}: it removes instruction-level acceleration from our convolution operator.
    \item \textbf{``DTMM-w/o-ins-vec''}: it completely disables our convolution operator and uses scalar instructions instead.
\end{itemize}

Figure~\ref{fig:eval_operator_ablation} (a) shows that without instruction-level acceleration, ``DTMM-w/o-ins'' increases model execution latency by 20.8--55.3\%. Also, our original convolution operator design plays a more important role, causing 231.8--635.0\% increased latency if it is disabled. This experiment shows the effectiveness of our proposed techniques in speeding up the execution of \systemname pruned models.

\parahead{Lane number of SIMD instructions} This is an important factor affecting the efficiency of our convolution operator design, as it affects how many dot products in convolution can be performed in parallel. In Figure~\ref{fig:eval_operator_ablation} (b), we measure the latency of executing five \systemname pruned layers of different sizes by decreasing the number of lanes from 16 to 4. The results show that latency increases by an average of 16.2\% and 48.9\% when the number of lanes is reduced to 8 and 4, respectively. In Figure~\ref{fig:eval_operator_ablation} (b), we use convolution layers instead of the whole model, since the whole model may contain unpruned layer(s) that are not executed by our operator.

Driven by the success of artificial intelligence, MCU processor manufactures begin to introduce SIMD with more lanes in their products. For example, the latest Cortex-M55 supports 16-lane SIMD MAC, while the second latest high-performance model Cortex-M7 released in 2014 only supports 2-lane SIMD MAC. For advanced processors with more lanes, \systemname can obtain higher speedup gains in the future.

\begin{figure}[h]
\centering
\vspace{-.1in}
\includegraphics[width=0.4\textwidth]{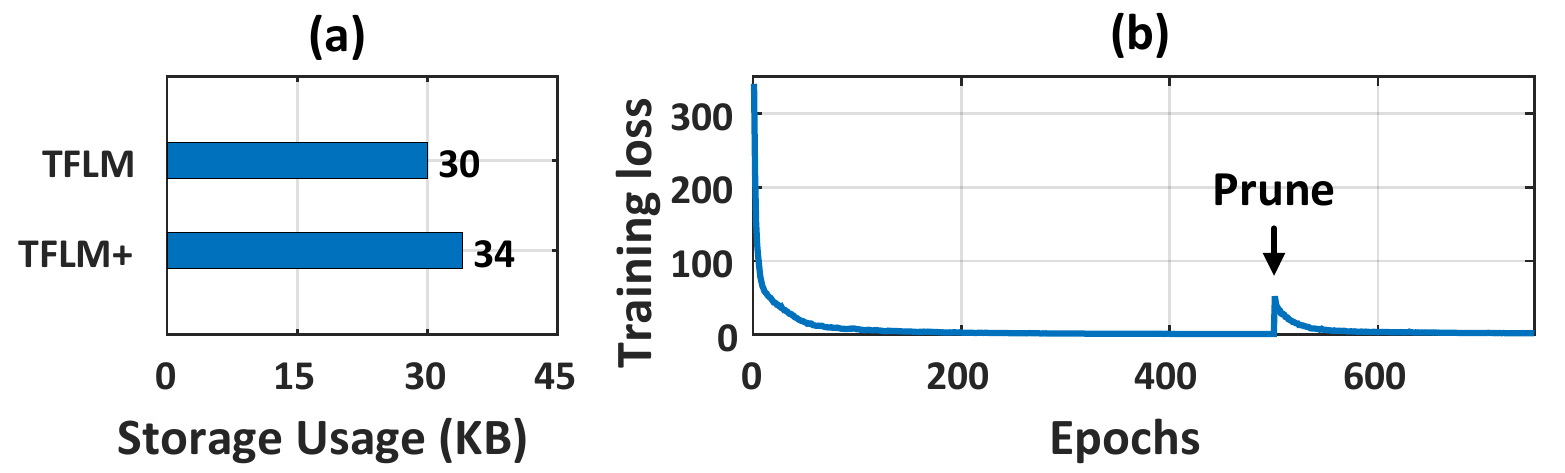}
\vspace{-.1in}
\caption{Overhead of (a) storage and (b) training time.}
\label{fig:eval_overhead} 
\vspace{-.1in}
\end{figure}

\subsection{System Overhead} 
\label{sec:eval:overhead}

\parahead{Storage overhead} When we include only necessary operators for the ML models used in the experiments, the size of the original Tensorflow Lite Micro (TFLM) is 30 KB. In \systemname, we further integrate our new operator, and its storage overhead is small, \eg, the size of TFLM with \systemname (TFLM+) only increases by 4 KB, as shown in Figure~\ref{fig:eval_overhead}(a).

\parahead{Training time} Finally, we study the training behavior of the model before and after pruning. In Figure~\ref{fig:eval_overhead}(b), we train the original YOLO for 500 epochs. We then use \systemname to prune and fine-tune the pruned model, which takes about half of the initial training time to converge again. 


\section{Related Works}
\label{sec:related}

\parahead{ML models on MCUs} Deploying ML models on MCUs leads to an emerging field TinyML~\cite{banbury2021micronets, David2021TensorFlowLM,warden2019tinyml} for useful applications, such as autonomous nano drone~\cite{duisterhof2019learning} and smart health bracelet~\cite{fyntanidou2020iot}. To support efficient model inference on MCUs, CMSIS-NN~\cite{Lai2018CMSISNNEN}, microTVM~\cite{chen2018tvm}, TinyEngine~\cite{lin2020mcunet} and TensorFlow Lite Micro~\cite{David2021TensorFlowLM} are proposed. Based on these commercial frameworks, recent works improve the performance of ML models on MCUs by recording operator execution to reduce memory consumption~\cite{liberis2019neural}, intermittent model execution~\cite{gobieski2019intelligence}, model architecture search~\cite{fedorov2019sparse}, etc. \systemname is a systematic solution to enable efficient local inference on MCUs.

\parahead{Model compression} There are existing studies to compress ML models on MCUs including knowledge distillation~\cite{Hinton2015DistillingTK}, low-rank factorization~\cite{bhattacharya2016sparsification}, and quantization~\cite{Han2016DeepCC}. They are orthogonal to \systemname and can be used simultaneously. For example, 8-bit quantization is currently used in \systemname (\S\ref{sec:impl}). 

Another popular model compression technique on MCUs is pruning~\cite{Han2016DeepCC,sui2021chip}, which can be divided into structured and unstructured methods. Structured pruning removes model weights according to a given structure~\cite{chen2018shallowing, Li2017PruningFF,he2017channel,wen2016learning}, which might significantly reduce the model accuracy when the compression ratio is high on the MCU. Unstructured pruning~\cite{niu2020patdnn,Han2016DeepCC} can retain the accuracy. However, models after pruning perform slowly due to the large overhead~\cite{ma2021non}. Some research~\cite{meng2020pruning} proposes to group weights to form new pruning units, which can have different forms by applying different constraints~\cite{niu2020patdnn,wen2016learning}. Unlike existing works above that focus on the structural design of pruning units, DTMM can actually run ML models with discrete weights after pruning on MCUs.

\parahead{Other methods} In addition to model compression, there have been works proposed to deploy ML models on MCUs from other perspectives~\cite{kwon2022yono,miao2021enabling}, where computation offloading is a typical example~\cite{huang2022real}. However, offloading relies on extra infrastructure for support, such as edge or cloud, which is not conductive to large-scale deployment. In \systemname, we overcome the resource constraints of weak IoT devices to achieve fully autonomous operation and local inference.


\section{Conclusion}
\label{sec:conclusion}

This paper presents \systemname, a specialized library for deploying TinyML models on low-end IoT devices like MCUs with pruning. We choose a suitable pruning unit and propose a dedicated storage structure to achieve high compression ratios while maintaining model accuracy. We also design a new operator, compatible with commercial ML frameworks, to efficiently execute \systemname pruned models, co-designed with a scheduler to derive the optimal pruning strategy. Evaluations show remarkable gains compared to state-of-the-art methods.

\section*{Acknowledgement}

This work is supported by the GRF grant (CityU 11202623) from Hong Kong RGC. Corresponding author is Zhenjiang Li.

\clearpage

\bibliographystyle{abbrv}
\bibliography{ref}
\end{document}